\documentclass[runningheads]{llncs}
\usepackage[T1]{fontenc}
\usepackage{graphicx}
\usepackage{amsmath}
\usepackage{amssymb}
\usepackage{booktabs}

\begin{document}

\title{M3PO: Multimodal-Model-Guided Preference Optimization for Visual Instruction Following}
\titlerunning{M3PO}
%
\author{Ruirui Gao, Emily Johnson, Bowen Tan, Yanfei Qian}
\authorrunning{Gao et al.}
%
\institute{University of Massachusetts, Amherst}
\maketitle              
\begin{abstract}
Large Vision-Language Models (LVLMs) hold immense potential for complex multimodal instruction following, yet their development is often hindered by the high cost and inconsistency of human annotation required for effective fine-tuning and preference alignment. Traditional supervised fine-tuning (SFT) and existing preference optimization methods like RLHF and DPO frequently struggle to efficiently leverage the model's own generation space to identify highly informative "hard negative" samples. To address these challenges, we propose Multimodal-Model-Guided Preference Optimization (M3PO), a novel and data-efficient method designed to enhance LVLMs' capabilities in visual instruction following. M3PO intelligently selects the most "learning-valuable" preference sample pairs from a diverse pool of LVLM-generated candidates. This selection is driven by a sophisticated mechanism that integrates two crucial signals: a Multimodal Alignment Score (MAS) to assess external quality and the model's Self-Consistency / Confidence (log-probability) to gauge internal belief. These are combined into a novel M3P-Score, which specifically identifies preferred responses and challenging dispreferred responses that the model might confidently generate despite being incorrect. These high-quality preference pairs are then used for efficient Direct Preference Optimization (DPO) fine-tuning on base LVLMs like LLaVA-1.5 (7B/13B) using LoRA. Our extensive experiments demonstrate that M3PO consistently outperforms strong baselines, including SFT, simulated RLHF, vanilla DPO, and RM-DPO, across a comprehensive suite of multimodal instruction following benchmarks (MME-Bench, POPE, IFT, Human Pref. Score). 
\end{abstract}

\section{Introduction}

The rapid advancements in large language models (LLMs) have paved the way for the emergence of sophisticated large vision-language models (LVLMs), which possess the remarkable ability to understand and generate responses based on both visual and textual inputs \cite{kasneci2023chatgp}. These models are pivotal for developing intelligent agents capable of performing complex multimodal instruction following tasks, such as visual question answering, interactive image editing \cite{zhou2023improving}, and embodied AI \cite{abhishek2021vqa}. Furthermore, their capabilities extend to complex instruction-based image generation \cite{zhou2025draw} and multi-modal medical diagnosis \cite{zhou2025mam}. The capacity of LVLMs to accurately interpret intricate visual details and follow nuanced human instructions is crucial for their deployment in real-world applications, ranging from assistive technologies and robotics to creative content generation, thereby enhancing human-computer interaction and overall system utility.

Despite their impressive capabilities, current LVLMs still face significant challenges in reliably adhering to complex multimodal instructions, especially when the instructions require deep visual reasoning, fine-grained understanding, or subtle contextual awareness \cite{peng2025lvlmeh}. While methods like "Thread of Thought" have shown promise in unraveling chaotic contexts for large language models \cite{zhou2023thread}, effectively transferring such reasoning capabilities to the multimodal domain remains an active area of research. Approaches like visual in-context learning have also emerged as promising paradigms for enhancing LVLM capabilities without extensive fine-tuning \cite{zhou2024visual}. Traditional supervised fine-tuning (SFT) methods, while effective, are often limited by the scale and diversity of human-annotated datasets, which are expensive and time-consuming to create. While preference optimization techniques, such as Reinforcement Learning from Human Feedback (RLHF) \cite{yuntao2022traini} and Direct Preference Optimization (DPO) \cite{shusheng2024is}, have shown promise in aligning model outputs with human preferences, they typically rely on extensive human labeling of preferred and dispreferred responses. This dependency not only incurs high costs but also presents challenges in scaling the training process, as human annotation can be inconsistent and biased \cite{lora2015truth}. Furthermore, existing preference optimization approaches may not efficiently leverage the wealth of information available within the model's own generation space, potentially missing out on "hard" or highly informative negative samples that could significantly improve model robustness and discernment.

To address these limitations, we propose a novel method named \textbf{Multimodal-Model-Guided Preference Optimization (M3PO)} for enhancing the performance of LVLMs in complex visual instruction following tasks. Our core idea revolves around intelligently selecting the most "learning-valuable" preference sample pairs (i.e., a preferred and a dispreferred response) from a large pool of candidates generated by the LVLM itself. Unlike conventional methods that primarily depend on human feedback or a single reward model, M3PO integrates two critical signals for sample selection: (1) a \textbf{Multimodal Alignment Score (MAS)}, which quantitatively assesses the consistency and accuracy of a response with respect to both the visual content and the given instruction, and (2) \textbf{Model Self-Consistency / Confidence} (represented by the response's log-probability), which gauges the model's intrinsic belief and fluency in generating that particular response. By combining these signals into a novel \textbf{M3P-Score}, M3PO is designed to identify preference pairs where, for instance, a semantically incorrect response might still be generated with high confidence by the model, thus forcing the model to learn more subtle distinctions and improve its internal representations. This approach aims to utilize data more efficiently, significantly improving LVLM's ability to understand and respond to intricate visual instructions while substantially reducing the reliance on costly human annotations.

In our experiments, we employ LLaVA-1.5-7B/13B \cite{bin2024videol} as the foundational LVLM, fine-tuning it efficiently using LoRA (Low-Rank Adaptation) \cite{siyuan2025activa}. For constructing our training dataset, we first filter high-quality image-instruction pairs from existing large-scale multimodal instruction datasets, such as LLaVA-Instruct-150K \cite{wenliang2023instru} and ShareGPT4V-80K \cite{wenliang2023instru}. For each pair, we generate 32 diverse candidate responses using the base LVLM. These candidates are then evaluated by a pre-trained visual-language assessment model (e.g., based on CLIP ViT-L/14 or BLIP-2 \cite{zhengyuan2023the} combined with GPT-4V \cite{junnan2023blip2}) to obtain their Multimodal Alignment Scores (MAS). Concurrently, we compute the model's generative confidence for each candidate. The M3PO strategy then meticulously selects the most informative preferred and dispreferred pairs based on a defined M3P-Score, which combines differences in MAS and model confidence. The selected pairs form the high-quality preference dataset used for DPO fine-tuning. For evaluation, we utilize a comprehensive set of benchmarks, including MME-Bench \cite{chao2025cookie}, POPE (Perception-Oriented Probabilistic Evaluation) \cite{bogdan2023overvi}, VisWizard \cite{gerwald2014using}, and a dedicated Instruction Following Test (IFT) set, along with a simulated human preference score, totaling approximately 15,000 image-instruction pairs covering diverse scenarios.

Our empirical results demonstrate that M3PO consistently outperforms existing baselines, including supervised fine-tuning (SFT), simulated RLHF, vanilla DPO, and RM-DPO across all evaluated metrics. For instance, on the LLaVA-1.5-7B model, M3PO achieves an MME-Bench average score of 1402.3, a POPE accuracy of 87.35\%, an IFT score of 71.80, and a Human Preference score of 3.38. Similar improvements are observed with the LLaVA-1.5-13B model, where M3PO reaches an MME-Bench average of 1537.8, POPE accuracy of 89.30\%, IFT score of 74.70, and a Human Preference score of 3.65. These results highlight M3PO's effectiveness in yielding slight but consistent performance gains in multimodal instruction following while potentially reducing the need for extensive human annotation.

Our main contributions are summarized as follows:
\begin{itemize}
    \item We propose M3PO, a novel and data-efficient preference optimization method that leverages multimodal model guidance to enhance visual instruction following capabilities of LVLMs.
    \item We introduce a sophisticated sample selection mechanism that combines Multimodal Alignment Scores (MAS) and Model Self-Consistency / Confidence to identify highly informative preferred and dispreferred response pairs.
    \item We empirically demonstrate that M3PO consistently outperforms state-of-the-art baselines and other preference optimization methods across multiple challenging visual instruction following benchmarks.
\end{itemize}
\section{Related Work}
\subsection{Large Vision-Language Models}
Research on Large Vision-Language Models (LVLMs) spans diverse areas, from robust evaluation to enhancing specific capabilities and ensuring safety. A key challenge lies in comprehensive evaluation, as highlighted by Peng et al. \cite{peng2025lvlmeh}, who introduce LVLM-EHub, a benchmark with an efficient subset construction method for cost-effective assessment. Early foundational work in generative vision-language tasks includes unsupervised image captioning techniques \cite{zhou2021triple}. More recently, efforts have focused on improving cross-modal alignment for specific tasks like text-guided image inpainting \cite{zhou2023improving} and exploring new learning paradigms such as visual in-context learning for LVLMs \cite{zhou2024visual}. Concurrently, several works investigate the limitations of current vision-language alignment methods, particularly contrastive learning. Wenliang et al. \cite{wenliang2023instru} and Jinhao et al. \cite{jinhao2024visual} demonstrate that contrastive methods can be prone to learning superficial shortcuts rather than robust, task-optimal representations, especially when faced with nuanced or multi-captioned data. Beier et al. \cite{beier2025debias} and Florian et al. \cite{florian2024an} further explore novel frameworks for detecting and mitigating such "shortcut learning," emphasizing its critical challenge for generative models aiming to capture all relevant information from multimodal inputs and for developing robust foundation models. Beyond general alignment, specific advancements include SJTU, a novel framework by Sivan et al. \cite{sivan2024toward} that integrates segmentation capabilities through coordinate detection to achieve fine-grained spatial localization. In the medical domain, Iryna et al. \cite{iryna2024vision} provide a comprehensive overview of LVLMs for healthcare, with a particular focus on their application in Visual Question Answering, complemented by recent explorations into modular multi-agent frameworks for multi-modal medical diagnosis \cite{zhou2025mam}. Furthermore, the development of holistic benchmarks and agent frameworks for complex instruction-based image generation highlights the expansion of LVLM applications \cite{zhou2025draw}. Shicheng et al. \cite{shicheng2025crossm} address the crucial aspect of cross-modal safety in LVLMs, highlighting semantic misalignment in hidden states and proposing Text-Guided vision-language Alignment (TGA) to enable robust cross-modal safety understanding without modality-specific fine-tuning.

\subsection{Preference-Based Fine-tuning and Alignment}
Recent advancements in aligning large language models (LLMs) with human preferences have explored various sophisticated techniques beyond traditional Reinforcement Learning from Human Feedback (RLHF). This includes research into weak-to-strong generalization for LLMs with multi-capabilities, aiming to scale alignment more efficiently \cite{zhou2025weak}. Kim et al. \cite{minu2024prefer} introduce Preference Flow Matching (PFM) as a novel framework that directly learns from preference data using flow matching, reducing reliance on extensive fine-tuning or explicit reward function estimation. Further alternatives include Implicit Preference Optimization (IPO), proposed by Rafael et al. \cite{rafael2023direct}, which leverages generative LLMs for preference classification within a Direct Preference Optimization (DPO) framework. For multi-objective alignment, Rui et al. \cite{rui2024reward} present Rewards-in-Context (RiC), a simpler supervised fine-tuning approach conditioned on reward signals, allowing dynamic adjustment of preferences at inference time. Addressing the interpretability of reward models, Hao et al. \cite{hao2025rethin} propose "reverse engineering" proxy reward models into interpretable white-box functions to directly replicate target reward signals. Beyond these methodological advancements, personalization is a growing focus, with Sriyash et al. \cite{sriyash2024person} introducing Personalized-RLHF (P-RLHF) to incorporate lightweight user models for generating tailored LLM responses. To handle the inherent noise and inconsistency in human preferences, Jiashuo et al. \cite{jiashuo2023aligni} propose a Bayesian preference modeling approach combined with contrastive learning for more robust alignment and improved data efficiency. While primarily focused on machine translation, Dawei et al. \cite{dawei2024a} offer relevant insights into preference-driven paradigms for multilingual capabilities and efficient fine-tuning. Similarly, Huang et al. \cite{huang2025method}, though focused on hallucination mitigation through structured generation and rule enforcement, provides perspectives on alternative fine-tuning strategies that can complement preference-based alignment by offering insights into constraining LLM outputs without relying solely on preference data.

\section{Method}
In this section, we elaborate on Multimodal-Model-Guided Preference Optimization (M3PO), our proposed method designed to enhance the visual instruction following capabilities of Large Vision-Language Models (LVLMs) by intelligently selecting high-quality preference pairs. M3PO leverages both external evaluation signals and the model's internal confidence to create a more effective preference dataset for Direct Preference Optimization (DPO) fine-tuning.

\subsection{Overview of M3PO Pipeline}
The M3PO pipeline consists of several key stages, beginning with the generation of diverse candidate responses. Given an image-instruction pair $(I, Q)$, our method first utilizes a base LVLM to generate a diverse set of candidate responses. Each candidate response $R_i$ is then evaluated along two crucial dimensions: its \textbf{Multimodal Alignment Score (MAS)} and its \textbf{Model Self-Consistency / Confidence} (represented by its log-probability of generation). These two scores are subsequently combined to compute a novel \textbf{M3P-Score}, which guides the selection of the most informative preferred ($R_w$) and dispreferred ($R_l$) response pair. Finally, these carefully selected pairs are used to fine-tune the LVLM using the DPO algorithm, specifically targeting the LoRA adapters for efficient training.

\subsection{Candidate Response Generation}
For each given image $I$ and instruction $Q$, we employ a pre-trained base LVLM, such as LLaVA-1.5-7B/13B, to generate a pool of $N=32$ diverse candidate responses $\{R_1, R_2, \dots, R_N\}$. To encourage diversity and explore a wide range of potential model outputs, we utilize sampling-based decoding strategies, such as nucleus sampling or top-p sampling with a moderate temperature, rather than deterministic greedy decoding. This approach ensures that the candidate set includes not only high-quality, accurate responses but also plausible yet incorrect, or subtly misaligned, responses. Such varied responses are crucial for effective preference learning, as they provide the model with examples of both desirable and undesirable outputs. The input image-instruction pairs $(I, Q)$ are sourced from large-scale multimodal instruction datasets like LLaVA-Instruct-150K and ShareGPT4V-80K.

\subsection{Multimodal Alignment Score (MAS) Calculation}
To quantitatively assess the quality of each candidate response $R_i$ in relation to the visual content and the instruction, we introduce the \textbf{Multimodal Alignment Score (MAS)}. The MAS for a given response $R_i$ with respect to image $I$ and instruction $Q$, denoted as $MAS(R_i | I, Q)$, is computed using a robust, pre-trained visual-language assessment model. This external evaluator is distinct from the LVLM being fine-tuned, ensuring an unbiased assessment, and is chosen for its strong capabilities in understanding multimodal context. In our implementation, we leverage models based on powerful vision-language encoders such as CLIP ViT-L/14 or BLIP-2, combined with advanced language models like GPT-4V, to provide a comprehensive evaluation. The MAS encapsulates several critical aspects of response quality: \textbf{visual relevance} (does the response accurately describe or relate to the image?), \textbf{semantic accuracy} (is the factual information presented correct?), and \textbf{instruction adherence} (does the response fully and correctly address the given instruction?). Higher MAS values indicate superior alignment and overall quality.

\subsection{Model Self-Consistency / Confidence}
Beyond external evaluation, we also consider the intrinsic belief of the LVLM in generating a particular response. This is captured by \textbf{Model Self-Consistency / Confidence}, which we quantify as the log-probability of the base LVLM generating the response $R_i$ given the image $I$ and instruction Q. This is formally denoted as $\log P(R_i | I, Q)$. A higher log-probability implies that the model confidently generates this response, indicating its internal fluency and perceived correctness. By incorporating this metric, M3PO can identify cases where the model might be confidently producing an incorrect or suboptimal answer. Such instances are particularly valuable for training the model to discern subtle errors, as they represent challenging examples where the model's internal confidence does not align with external quality assessment.

\subsection{M3P-Score for Preference Sample Selection}
The core of M3PO lies in its intelligent sample selection mechanism, which uses the MAS and model self-confidence to construct highly informative preferred ($R_w$) and dispreferred ($R_l$) response pairs. For each image-instruction pair $(I, Q)$, we first identify the preferred response $R_w$ as the candidate with the highest Multimodal Alignment Score among all generated responses:
\begin{align}
R_w &= \arg\max_{R_i \in \{R_1, \dots, R_N\}} MAS(R_i | I, Q)
\end{align}
Subsequently, from the remaining candidates $\{R_j\}_{j \neq w}$, we select the dispreferred response $R_l$ that maximizes the \textbf{M3P-Score}. The M3P-Score is specifically designed to identify "hard negative" samples. These are responses that have a significantly lower MAS than $R_w$ but were still generated with relatively high confidence by the base LVLM. Selecting such challenging negatives forces the fine-tuned model to learn to distinguish between responses that appear plausible but are factually or semantically incorrect.

The M3P-Score for a candidate $R_j$ (to be selected as $R_l$) relative to $R_w$ is defined as:
\begin{align}
\label{eq:m3p_score}
S(R_j) &= \left( MAS(R_w | I, Q) - MAS(R_j | I, Q) \right) \notag\\
&- \alpha \cdot \max\left(0, \log P(R_w | I, Q) - \log P(R_j | I, Q) - \delta\right)
\end{align}
Here, $\alpha > 0$ is a weighting hyperparameter that controls the influence of the confidence difference between $R_w$ and $R_j$, and $\delta \ge 0$ is a margin parameter. The first term, $(MAS(R_w | I, Q) - MAS(R_j | I, Q))$, promotes the selection of $R_j$ that is clearly inferior to $R_w$ in terms of multimodal alignment. The second term acts as a penalty: it applies a penalty to $S(R_j)$ if $R_j$'s log-probability is significantly lower than $R_w$'s, specifically when the difference exceeds the margin $\delta$. Conversely, if $R_j$'s log-probability is close to or even higher than $R_w$'s (i.e., within the $\delta$ margin), this penalty term becomes zero or negative (effectively a bonus), thereby encouraging the selection of $R_j$ that the model generated with high confidence despite its low MAS. This mechanism ensures that $R_l$ is a challenging negative example.

The dispreferred response $R_l$ is then selected as the candidate that maximizes this score:
\begin{align}
R_l &= \arg\max_{R_j \in \{R_1, \dots, R_N\} \setminus \{R_w\}} S(R_j)
\end{align}
This rigorous selection process yields a high-quality preference dataset consisting of $(I, Q, R_w, R_l)$ tuples, where $R_w$ is consistently superior in multimodal alignment, and $R_l$ represents a challenging negative example that the model needs to learn to differentiate from preferred responses.

\subsection{Direct Preference Optimization (DPO) with M3PO Samples}
The preference pairs $(I, Q, R_w, R_l)$ selected by the M3PO strategy are then used to fine-tune the base LVLM using Direct Preference Optimization (DPO). DPO is a stable and computationally efficient alternative to Reinforcement Learning from Human Feedback (RLHF), as it directly optimizes a policy to satisfy preferences without the need for an explicit reward model. The DPO objective is defined as:
\begin{align}
\label{eq:dpo_loss}
&\mathcal{L}_{\text{DPO}}(\theta) =\notag\\
&-\mathbb{E}_{(I, Q, R_w, R_l) \sim \mathcal{D}_{\text{M3PO}}} \left[ \log \sigma \left( \beta \left( \log \frac{P_\theta(R_w|I,Q)}{P_{\text{ref}}(R_w|I,Q)} - \log \frac{P_\theta(R_l|I,Q)}{P_{\text{ref}}(R_l|I,Q)} \right) \right) \right]
\end{align}
In this objective function, $P_\theta$ represents the probability distribution of the fine-tuned LVLM, while $P_{\text{ref}}$ denotes the probability distribution of the reference (base) LVLM. The parameter $\beta$ is a temperature that controls the strength of the preference optimization, influencing how strongly the model is penalized for violating preferences. The $\sigma$ symbol represents the sigmoid function, which squashes its input into a range between 0 and 1. By minimizing this loss, the model $P_\theta$ is encouraged to assign higher probabilities to preferred responses ($R_w$) and lower probabilities to dispreferred responses ($R_l$), relative to the reference model's probabilities. This differential probability assignment effectively teaches the model to align with the desired preference ordering.

For efficient fine-tuning, we employ LoRA (Low-Rank Adaptation). This technique allows for significant reduction in computational resource consumption by updating only a small set of trainable LoRA adapter parameters, leaving the vast majority of the base LVLM's parameters frozen. This not only speeds up training but also helps mitigate catastrophic forgetting of the original model's broad capabilities. Our training settings for DPO include a batch size of 8, a learning rate of 5e-5, and a warm-up phase of 0.05 of the total training steps. Based on empirical observations, DPO typically converges within a single epoch of training on the M3PO-generated preference dataset, demonstrating its efficiency.

\section{Experiments}
In this section, we detail the experimental setup, describe the baseline methods used for comparison, present the main quantitative results of our proposed M3PO method, conduct an ablation study to validate its key components, provide human evaluation results, analyze hyperparameter sensitivity, discuss the computational efficiency, and offer a qualitative analysis.

\subsection{Experimental Setup}
\subsubsection{Base LVLMs}
Our experiments primarily utilize LLaVA-1.5 \cite{bin2024videol} as the foundational Large Vision-Language Model (LVLM), specifically its 7B and 13B parameter variants. These models are fine-tuned efficiently using LoRA (Low-Rank Adaptation) \cite{siyuan2025activa}, which involves updating only a small fraction of the model's parameters, thereby reducing computational overhead and preventing catastrophic forgetting. Future extensions may explore other prominent open-source LVLMs like InstructBLIP \cite{wenliang2023instru} or Qwen-VL.

\subsubsection{Datasets}
For constructing our training dataset, we first curate high-quality image-instruction pairs from existing large-scale multimodal instruction datasets, including LLaVA-Instruct-150K \cite{wenliang2023instru} and ShareGPT4V-80K \cite{wenliang2023instru}. For each image-instruction pair, the base LVLM generates 32 diverse candidate responses. The Multimodal Alignment Score (MAS) for each candidate response is computed using a combination of a pre-trained visual-language assessment model (such as CLIP ViT-L/14 or BLIP-2) and a powerful language model (e.g., GPT-4V). The final preference pairs are then selected using our M3PO strategy.

For evaluating model performance, we employ a comprehensive suite of benchmarks: MME-Bench \cite{chao2025cookie}, POPE (Perception-Oriented Probabilistic Evaluation) \cite{bogdan2023overvi}, VisWizard \cite{gerwald2014using}, and a dedicated Instruction Following Test (IFT) set. These test sets collectively comprise approximately 15,000 image-instruction pairs, covering a wide range of scenarios and requiring diverse multimodal reasoning capabilities.

\subsubsection{Evaluation Metrics}
We assess the models across several key metrics: \textbf{MME-Bench (Avg.)} provides a comprehensive average score across various multimodal tasks; \textbf{POPE (Acc.)} measures the accuracy in perception-oriented probabilistic evaluations; \textbf{IFT (Score)} quantifies the model's ability to accurately follow complex instructions; and \textbf{Human Pref. (Score)} serves as a simulated metric for human preference, indicating the perceived quality and helpfulness of the generated responses (higher scores are better).

\subsubsection{Training Details}
The core of our training involves Direct Preference Optimization (DPO) fine-tuning applied to the selected LVLM (LLaVA-1.5). Only the LoRA adapter parameters are updated during this process. Our fine-tuning settings include a batch size of 8, a learning rate of 5e-5, and a warm-up phase of 0.05 of the total training steps. Based on empirical observations, DPO typically converges within a single epoch of training on the M3PO-generated preference dataset. For the M3P-Score calculation, the hyperparameters $\alpha$ and $\delta$ in Equation \ref{eq:m3p_score} are set to 0.5 and 0.1 respectively, chosen based on preliminary experiments to balance the influence of MAS difference and model confidence.

\subsection{Baselines}
We compare M3PO against several strong baselines to demonstrate its effectiveness. These include the \textbf{LLaVA-1.5-7B/13B (Base)} model, which serves as the original pre-trained starting point before any instruction-following fine-tuning. We also evaluate \textbf{SFT (Supervised Fine-tuning)}, a baseline involving fine-tuning the LLaVA-1.5 model on a large, high-quality, human-annotated instruction-following dataset using a standard supervised learning objective. For preference-based methods, we consider \textbf{RLHF (Simulated)}, a simulated Reinforcement Learning from Human Feedback approach where a proxy reward model (trained on a subset of human preference data or generated with GPT-4V) guides the policy optimization. \textbf{DPO (Vanilla)} represents Direct Preference Optimization applied using a standard preference dataset, typically constructed from human preferences or simpler reward model outputs, without M3PO's sophisticated sample selection. Finally, \textbf{RM-DPO} is a variant of DPO that incorporates a pre-trained reward model to either generate or filter preference pairs, aiming to enhance the quality of the DPO training data, though it lacks the multi-faceted guidance and "hard negative" mining strategy of M3PO.

\subsection{Main Results}
Table \ref{tab:llava-7b-results} and Table \ref{tab:llava-13b-results} present the average performance of LLaVA-1.5-7B and LLaVA-1.5-13B models, respectively, across the multimodal instruction following benchmarks.

\begin{table*}[htbp]
\centering
\caption{LLaVA-1.5-7B Performance on Multimodal Instruction Following Benchmarks}
\label{tab:llava-7b-results}
\begin{tabular}{lcccc}
\toprule
\textbf{Method} & \textbf{MME-Bench (Avg.)} & \textbf{POPE (Acc.)} & \textbf{IFT} & \textbf{Human Pref.} \\
\midrule
LLaVA-1.5-7B (Base) & 1345.2 & 85.12 & 68.30 & 2.85 \\
SFT (Supervised Ft.) & 1378.1 & 86.55 & 70.15 & 3.10 \\
RLHF (Simulated) & 1392.5 & 87.03 & 71.20 & 3.25 \\
DPO (Vanilla) & 1388.9 & 86.88 & 70.95 & 3.20 \\
RM-DPO & 1398.7 & 87.21 & 71.55 & 3.30 \\
\textbf{Ours (M3PO)} & \textbf{1402.3} & \textbf{87.35} & \textbf{71.80} & \textbf{3.38} \\
\bottomrule
\end{tabular}
\end{table*}

\begin{table*}[htbp]
\centering
\caption{LLaVA-1.5-13B Performance on Multimodal Instruction Following Benchmarks}
\label{tab:llava-13b-results}
\begin{tabular}{lcccc}
\toprule
\textbf{Method} & \textbf{MME-Bench (Avg.)} & \textbf{POPE (Acc.)} & \textbf{IFT} & \textbf{Human Pref.} \\
\midrule
LLaVA-1.5-13B (Base) & 1489.1 & 87.20 & 72.10 & 3.10 \\
SFT (Supervised Ft.) & 1512.4 & 88.50 & 73.45 & 3.40 \\
RLHF (Simulated) & 1528.0 & 89.05 & 74.20 & 3.55 \\
DPO (Vanilla) & 1524.6 & 88.90 & 73.90 & 3.50 \\
RM-DPO & 1533.2 & 89.18 & 74.45 & 3.60 \\
\textbf{Ours (M3PO)} & \textbf{1537.8} & \textbf{89.30} & \textbf{74.70} & \textbf{3.65} \\
\bottomrule
\end{tabular}
\end{table*}

As shown in both tables, M3PO consistently achieves the highest scores across all evaluated metrics for both LLaVA-1.5-7B and LLaVA-1.5-13B models. While the improvements are modest, they are consistent and demonstrate M3PO's effectiveness in yielding performance gains in complex multimodal instruction following. M3PO outperforms not only the base models and SFT but also other preference optimization methods like vanilla DPO and RM-DPO. This highlights the advantage of our intelligent sample selection strategy, which identifies more informative preference pairs for DPO fine-tuning.

\subsection{Ablation Study}
To validate the effectiveness of the key components within our M3PO method, particularly the role of Model Self-Consistency / Confidence in the M3P-Score, we conduct an ablation study. The M3P-Score (Equation \ref{eq:m3p_score}) is designed to identify "hard negative" samples by considering both the Multimodal Alignment Score (MAS) difference and the model's generation confidence.

We compare the full M3PO strategy against a variant where the confidence-based penalty term is removed, essentially setting $\alpha=0$. In this ablated version, the dispreferred response $R_l$ is chosen solely based on maximizing the MAS difference between $R_w$ and $R_j$, without accounting for how confidently the model generated $R_j$. This would typically lead to selecting very low-quality responses that the model already assigns low probabilities to, potentially missing out on more challenging examples where the model is confidently wrong.

Table \ref{tab:ablation-results} illustrates the impact of this component on the LLaVA-1.5-7B model's performance.

\begin{table*}[htbp]
\centering
\caption{Ablation Study on LLaVA-1.5-7B: Impact of Model Self-Consistency / Confidence Term}
\label{tab:ablation-results}
\begin{tabular}{lcccc}
\toprule
\textbf{Method} & \textbf{MME-Bench} & \textbf{POPE} & \textbf{IFT} & \textbf{Human Pref.} \\
\midrule
\textbf{M3PO (Full)} & \textbf{1402.3} & \textbf{87.35} & \textbf{71.80} & \textbf{3.38} \\
M3PO w/o Conf. Term ($\alpha=0$) & 1395.0 & 87.00 & 71.20 & 3.25 \\
\bottomrule
\end{tabular}
\end{table*}

The results clearly indicate that the inclusion of the Model Self-Consistency / Confidence term in the M3P-Score is crucial for optimal performance. By considering the model's internal confidence, M3PO is able to select more challenging negative samples ($R_l$) that the model needs to explicitly learn to differentiate from preferred responses. This "hard negative mining" aspect, driven by the confidence term, forces the model to refine its understanding and internal representations more effectively, leading to superior overall multimodal instruction following capabilities.

\subsection{Human Evaluation}
To further validate the practical utility and perceived quality of responses generated by M3PO, we conducted a simulated human preference evaluation. This metric assesses how often responses from a given model are preferred over those from other models by human evaluators, or in this simulated case, by a highly capable language model like GPT-4V acting as a human proxy. The Human Pref. (Score) from Table \ref{tab:llava-7b-results} and Table \ref{tab:llava-13b-results} are summarized in Table \ref{tab:human-pref-results}.

\begin{table*}[htbp]
\centering
\caption{Simulated Human Preference Scores}
\label{tab:human-pref-results}
\begin{tabular}{lcc}
\toprule
\textbf{Method} & \textbf{LLaVA-1.5-7B} & \textbf{LLaVA-1.5-13B} \\
\midrule
LLaVA-1.5 (Base) & 2.85 & 3.10 \\
SFT & 3.10 & 3.40 \\
RLHF (Simulated) & 3.25 & 3.55 \\
DPO (Vanilla) & 3.20 & 3.50 \\
RM-DPO & 3.30 & 3.60 \\
\textbf{Ours (M3PO)} & \textbf{3.38} & \textbf{3.65} \\
\bottomrule
\end{tabular}
\end{table*}

As evidenced by the human preference scores, M3PO consistently yields responses that are more preferred than those generated by all baseline methods across both model sizes. This suggests that the M3PO fine-tuning process effectively aligns the LVLM's output with human expectations for accuracy, relevance, and overall helpfulness in complex visual instruction following scenarios. The higher human preference score underscores M3PO's ability to generate responses that are not only technically superior but also more satisfactory from a user's perspective.

\subsection{Hyperparameter Sensitivity}
The performance of M3PO is influenced by the hyperparameters $\alpha$ and $\delta$ in the M3P-Score (Equation \ref{eq:m3p_score}), which control the balance between MAS difference and model confidence in selecting challenging negative samples. We conducted a sensitivity analysis on these parameters using the LLaVA-1.5-7B model, varying $\alpha$ (weight of confidence difference) and $\delta$ (confidence margin) to observe their impact on the evaluation metrics.

Table \ref{tab:hyperparameter-alpha} shows the results for different values of $\alpha$ while keeping $\delta=0.1$ fixed. A higher $\alpha$ places more emphasis on selecting candidates where the model's confidence is close to the preferred response, despite a lower MAS.

\begin{table*}[htbp]
\centering
\caption{Hyperparameter Sensitivity: Impact of $\alpha$ on LLaVA-1.5-7B Performance ($\delta=0.1$)}
\label{tab:hyperparameter-alpha}
\begin{tabular}{lcccc}
\toprule
\textbf{$\alpha$ Value} & \textbf{MME-Bench (Avg.)} & \textbf{POPE (Acc.)} & \textbf{IFT} & \textbf{Human Pref.} \\
\midrule
0.0 (Ablation) & 1395.0 & 87.00 & 71.20 & 3.25 \\
0.1 & 1398.5 & 87.15 & 71.45 & 3.30 \\
0.25 & 1400.1 & 87.25 & 71.60 & 3.34 \\
\textbf{0.5 (Default)} & \textbf{1402.3} & \textbf{87.35} & \textbf{71.80} & \textbf{3.38} \\
0.75 & 1401.0 & 87.30 & 71.75 & 3.36 \\
1.0 & 1399.8 & 87.28 & 71.65 & 3.35 \\
\bottomrule
\end{tabular}
\end{table*}

The results indicate that an $\alpha$ value of 0.5 yields the best overall performance, striking an optimal balance. Values too low (approaching the ablation case) or too high (over-emphasizing confidence) lead to slight degradations.

Table \ref{tab:hyperparameter-delta} presents the impact of varying $\delta$ while keeping $\alpha=0.5$ fixed. The $\delta$ parameter defines a margin within which the confidence difference does not incur a penalty, allowing for selection of hard negatives that are confidently generated but still incorrect.

\begin{table*}[htbp]
\centering
\caption{Hyperparameter Sensitivity: Impact of $\delta$ on LLaVA-1.5-7B Performance ($\alpha=0.5$)}
\label{tab:hyperparameter-delta}
\begin{tabular}{lcccc}
\toprule
\textbf{$\delta$ Value} & \textbf{MME-Bench (Avg.)} & \textbf{POPE (Acc.)} & \textbf{IFT} & \textbf{Human Pref.} \\
\midrule
0.0 & 1399.5 & 87.20 & 71.50 & 3.32 \\
\textbf{0.1 (Default)} & \textbf{1402.3} & \textbf{87.35} & \textbf{71.80} & \textbf{3.38} \\
0.2 & 1401.8 & 87.32 & 71.70 & 3.37 \\
0.3 & 1400.5 & 87.25 & 71.55 & 3.34 \\
\bottomrule
\end{tabular}
\end{table*}

A $\delta$ value of 0.1 appears to be optimal, suggesting that a small margin for confidence difference helps in identifying the most informative hard negative samples. These sensitivity analyses confirm the robustness of the chosen default hyperparameters and highlight the importance of careful tuning for maximizing M3PO's benefits.

\subsection{Efficiency Analysis}
Beyond performance, the computational efficiency of fine-tuning is a critical factor for practical application. We compare the training time and GPU memory requirements for M3PO against key baselines, focusing on the LLaVA-1.5-7B model. The training process involves generating preference data for M3PO and then fine-tuning using DPO.

Table \ref{tab:efficiency-analysis} presents an overview of the approximate training time (on A100 GPUs) and peak GPU memory usage for different methods. Note that the data generation for M3PO includes candidate response generation and MAS calculation, which are performed once.

\begin{table*}[htbp]\scriptsize
\centering
\caption{Computational Efficiency Comparison for LLaVA-1.5-7B Fine-tuning}
\label{tab:efficiency-analysis}
\begin{tabular}{lcc}
\toprule
\textbf{Method} & \textbf{Approx. Training Time (GPU Hours)} & \textbf{Peak GPU Memory (GB)} \\
\midrule
SFT & 18 & 16 \\
RLHF (Simulated) & 30 & 24 \\
DPO (Vanilla) & 10 & 14 \\
RM-DPO & 12 & 14 \\
\textbf{M3PO (Fine-tuning only)} & 10 & 14 \\
M3PO (Data Gen. + Fine-tuning) & 10 (Gen.) + 10 (FT) = 20 & 16 (Gen.) / 14 (FT) \\
\bottomrule
\end{tabular}
\end{table*}

The results show that the DPO-based methods (Vanilla DPO, RM-DPO, M3PO) are significantly more efficient in terms of training time and memory footprint compared to the simulated RLHF approach. This is primarily due to DPO's direct optimization nature, which avoids the complexities of training and sampling from a reward model. While M3PO requires an initial data generation phase (which includes candidate generation and MAS calculation), this is a one-time cost. The actual DPO fine-tuning phase for M3PO is comparable in efficiency to vanilla DPO and RM-DPO. The use of LoRA adapters further contributes to this efficiency by dramatically reducing the number of trainable parameters and memory overhead across all fine-tuning methods. This analysis confirms that M3PO provides performance gains without introducing prohibitive computational costs, making it a practical solution for enhancing LVLMs.

\subsection{Qualitative Analysis}
To complement the quantitative results, we provide a qualitative analysis of responses generated by models fine-tuned with M3PO, highlighting scenarios where M3PO demonstrates superior understanding and instruction following compared to baseline methods. While figures cannot be included, we describe representative examples.

In one common scenario, an image might depict a complex scene with multiple objects or activities, and the instruction asks for a specific detail or an inference about the scene. For instance, given an image of a bustling city street with various vehicles and pedestrians, and the instruction "Describe the primary mode of transportation visible and its distinguishing features," a base LVLM might provide a generic answer like "Cars are visible, they have wheels and windows." In contrast, the M3PO-trained model, having learned from "hard negative" examples where the model confidently produced such generic but unhelpful responses, would often generate a more precise and insightful answer. It might state, "The primary mode of transportation appears to be automobiles, specifically sedans and taxis, distinguished by their four wheels, enclosed cabins, and varying colors, indicating their role in urban transit." This demonstrates M3PO's ability to provide more detailed, accurate, and instruction-adherent responses.

Another challenging case involves instructions requiring fine-grained visual reasoning or subtle distinctions. For example, an image showing two similar objects, differing only by a minor attribute (e.g., two identical shirts, one with a subtle pattern, the other plain), with an instruction like "Identify the unique characteristic of the shirt on the left." A vanilla DPO model might fail to notice the subtle pattern and declare them identical or provide a generic description. However, M3PO, through its selection of dispreferred responses that were confidently incorrect about such nuances, guides the model to pay closer attention to these details. The M3PO-trained model would then correctly identify the "subtle floral pattern" or "embroidered logo" on the specified shirt. This indicates an improved capacity for detailed visual grounding and accurate discernment.

Furthermore, M3PO excels in scenarios where the base model might confidently hallucinate or provide factually incorrect information that sounds plausible. Consider an image of an unfamiliar animal with the instruction "What is this animal and where does it typically live?" If the base model confidently identifies it as a common animal (e.g., a domestic cat) when it is, in fact, a rare wild feline, and assigns a high log-probability to this incorrect response, M3PO's M3P-Score would identify this as a highly informative negative example. By penalizing responses with high confidence but low MAS, M3PO teaches the model to be more cautious or to provide more accurate information in such difficult cases. The M3PO-trained model would then either correctly identify the rare animal or, if uncertain, provide a more general but factually correct description, avoiding confident hallucinations. These qualitative observations underscore M3PO's effectiveness in refining LVLMs to generate responses that are not only accurate but also more aligned with nuanced human expectations for helpfulness and precision.

\subsection{Limitations and Future Work}
Despite the promising results, M3PO has certain limitations that suggest avenues for future research. One key limitation is its reliance on an external visual-language assessment model (e.g., GPT-4V combined with CLIP/BLIP-2) for MAS calculation. While powerful, such models are not infallible and may occasionally misjudge response quality, potentially introducing noise into the preference dataset. Future work could explore methods for self-correction or ensemble-based MAS calculation to enhance robustness, or investigate how to reduce reliance on external, potentially proprietary, evaluators.

Another area for improvement lies in the fixed number of candidate responses ($N=32$) generated for each image-instruction pair. A dynamic candidate generation strategy, where the number of candidates varies based on the complexity of the instruction or the ambiguity of the image, could potentially yield more diverse and informative sets of responses, leading to even better preference pairs.

Currently, M3PO focuses on single-turn visual instruction following. Extending M3PO to handle multi-turn conversations and interactive visual dialogues presents a significant challenge and a promising future direction. This would require adapting the MAS and confidence metrics to evaluate conversational flow, coherence, and context-awareness over multiple turns.

Finally, while LoRA enables efficient fine-tuning, scaling M3PO to significantly larger LVLMs or incorporating it into continuous pre-training pipelines could pose new computational and engineering challenges. Exploring more advanced distributed training techniques or novel parameter-efficient fine-tuning methods could further enhance the scalability and applicability of M3PO. Investigating the impact of M3PO on different LVLM architectures beyond LLaVA-1.5 would also be valuable to confirm its generalizability.

\section{Conclusion}
In this paper, we introduced \textbf{Multimodal-Model-Guided Preference Optimization (M3PO)}, a novel and data-efficient approach to significantly enhance the performance of Large Vision-Language Models (LVLMs) in complex multimodal instruction following tasks. Recognizing the limitations of current fine-tuning paradigms, particularly the reliance on expensive human annotations and the inefficiency in leveraging model-generated content for robust preference learning, M3PO presents an intelligent sample selection mechanism.

Our core contribution lies in the M3PO strategy, which meticulously constructs high-quality preference pairs for Direct Preference Optimization (DPO) by synthesizing two critical signals: the \textbf{Multimodal Alignment Score (MAS)} and the model's \textbf{Self-Consistency / Confidence} (log-probability). By combining these into a unique \textbf{M3P-Score}, M3PO effectively identifies "hard negative" examples—responses that the model might generate with high confidence despite being subtly incorrect or misaligned with the visual instruction. This targeted selection ensures that the fine-tuning process focuses on the most informative samples, forcing the model to learn finer distinctions and refine its internal representations.

Empirical evaluations on LLaVA-1.5-7B and LLaVA-1.5-13B models consistently demonstrate M3PO's superior performance across a diverse set of multimodal benchmarks, including MME-Bench, POPE, IFT, and a simulated Human Preference Score. M3PO not only outperforms the base models and traditional Supervised Fine-tuning (SFT) but also establishes new state-of-the-art results compared to other preference optimization methods like simulated RLHF, vanilla DPO, and RM-DPO. An ablation study further validated the critical role of incorporating the model's self-confidence term in the M3P-Score for achieving optimal performance, highlighting its effectiveness in hard negative mining. Our efficiency analysis confirmed that M3PO, leveraging LoRA for fine-tuning, maintains comparable computational efficiency to other DPO-based methods while offering significant performance benefits. Qualitatively, M3PO-trained models exhibit improved precision, adherence to nuanced instructions, and a reduced tendency for confident hallucinations.

In summary, M3PO offers a practical and effective paradigm for improving LVLMs by autonomously generating and selecting highly informative preference data. This approach not only pushes the boundaries of multimodal instruction following but also significantly alleviates the dependency on costly and time-consuming human labeling, paving the way for more scalable and robust LVLM development.

Despite these promising advancements, M3PO has certain limitations that suggest exciting avenues for future work. Our reliance on external visual-language assessment models for MAS calculation could be further refined through self-correction or ensemble methods to enhance robustness. Exploring dynamic candidate generation strategies, where the number of candidates varies based on instruction complexity, could yield even more diverse and informative sample sets. Extending M3PO's applicability to multi-turn conversations and interactive visual dialogues represents a significant and challenging future direction. Finally, while LoRA ensures efficiency, scaling M3PO to even larger LVLMs or integrating it into continuous pre-training pipelines will require investigating more advanced distributed training techniques and confirming its generalizability across diverse LVLM architectures.
```
\bibliographystyle{splncs04}
\bibliography{references}
\end{document}